\documentclass[journal,twoside,web]{ieeecolor}
\usepackage{generic}
\usepackage{array}
\usepackage{cite}
\usepackage{amsmath,amssymb,amsfonts}
\usepackage{multirow}
\usepackage{makecell}
\usepackage{graphicx}
\usepackage{algorithm}
\usepackage{algpseudocode}
\usepackage{hyperref}
\hypersetup{hidelinks=true}
\usepackage{textcomp}
\def\BibTeX{{\rm B\kern-.05em{\sc i\kern-.025em b}\kern-.08em
    T\kern-.1667em\lower.7ex\hbox{E}\kern-.125emX}}
\begin{document}
\title{Enhancing Fairness in Skin Lesion
Classification for Medical Diagnosis Using Prune Learning}
\author{Kuniko Paxton, Koorosh Aslansefat, Dhavalkumar Thakker, Yiannis Papadopoulos and Tanaya Maslekar
\thanks{17 August 2025}
\thanks{Kuniko Paxton is with the School of Digital and Physical Sciences at the University of Hull, Cottingham Road, Hull, HU6 7RX, United Kingdom (e-mail: k.azuma-2021@hull.ac.uk).}
\thanks{Koorosh Aslansefat is with the School of Digital and Physical Sciences at the University of Hull, Cottingham Road, Hull, HU6 7RX, United Kingdom (e-mail: K.Aslansefat@hull.ac.uk).}
\thanks{Dhavalkumar Thakker is with the School of Digital and Physical Sciences at the University of Hull, Cottingham Road, Hull, HU6 7RX, United Kingdom (e-mail: D.Thakker@hull.ac.uk).}
\thanks{Yiannis Papadopoulos is with the School of Digital and Physical Sciences at the University of Hull, Cottingham Road, Hull, HU6 7RX, United Kingdom (e-mail: Y.I.Papadopoulos@hull.ac.uk).}
\thanks{Tanaya Maslekar is with Leeds Teaching Hospitals NHS Trust, Great George St, Leeds LS1 3EX, United Kingdom  (e-mail:tanaya.maslekar4@nhs.net).}}

\maketitle

\begin{abstract}
Recent advances in deep learning have significantly improved the accuracy of skin lesion classification models, supporting medical diagnoses and promoting equitable healthcare. However, concerns remain about potential biases related to skin color, which can impact diagnostic outcomes. Ensuring fairness is challenging due to difficulties in classifying skin tones, high computational demands, and the complexity of objectively verifying fairness. To address these challenges, we propose a fairness algorithm for skin lesion classification that overcomes the challenges associated with achieving diagnostic fairness across varying skin tones. By calculating the skewness of the feature map in the convolution layer of the VGG (Visual Geometry Group) network and the patches and the heads of the Vision Transformer, our method reduces unnecessary channels related to skin tone, focusing instead on the lesion area. This approach lowers computational costs and mitigates bias without relying on conventional statistical methods. It potentially reduces model size while maintaining fairness, making it more practical for real-world applications.
\end{abstract}


\section{Introduction}
In recent years, the predictive performance of deep learning models for skin lesion classification has improved significantly \cite{esteva2017dermatologist,brinker2019deep}. These models have the potential to assist medical professionals in diagnosing diseases more efficiently, allowing for timely medical diagnosis. Furthermore, they can enable mobile and web-based self-diagnosis tools, improving access to healthcare and promoting equitable delivery in low- and middle-income regions. However, despite these advances, concerns persist that algorithmic and data biases can compromise fairness in healthcare \cite{chinta2025ai,ueda2024fairness}. 
In this context, compromising fairness refers to systematic disparities in diagnostic accuracy or outcomes across patient groups. In particular, fairness concerns are heightened when models misdiagnose individuals based on protected attributes rooted in their historical and social backgrounds, such as skin color, thus eroding trust in their diagnostic outputs. These concerns are especially pronounced in the context of skin cancer detection \cite{kinyanjui2019estimating,alipour2024skin} because most training datasets are heavily biased toward lighter skin tones, leading to reduced diagnostic accuracy for people with darker skin. This imbalance can result in higher rates of misdiagnosis or delayed diagnosis, which directly impacts health outcomes. Furthermore, variations in skin pigmentation can affect the visual characteristics of lesions, making it more challenging for algorithms to generalize fairly across diverse populations if not explicitly accounted for. 

To address these issues, several academic studies have proposed fairness-enhancing methods in deep learning. Six notable technical approaches include: (1) regularization and reweighting, (2) minority image augmentation and generation, (3) counterfactual methods, (4) federated learning, (5) unlearning or adversarial approaches, and (6) prune learning. While each has demonstrated some capacity to reduce performance bias, all face practical challenges. These include difficulties in objectively defining and measuring skin color, trade-offs in model performance, high computational costs, and complexity of implementation. As \cite{sylvester2018applied} points out, real-world fairness solutions must be both effective and practical, rather than purely theoretical or overly complex.

For example, regularization \cite{pundhir2024biasing,e2022differentiable} and reweighting methods \cite{du2022fairdisco} attempt to calibrate imbalances across skin tones. FairDisCo (du2022fairdisco) employs a disentanglement contrastive learning approach that removes skin-tone information from learned representations while enhancing condition-relevant features. Pundhir et al introduce a two‑teacher knowledge transfer strategy, where each “biased” teacher is trained on different sensitive attributes, and their knowledge is combined into a student model using a weighted loss balancing biasing and debiasing terms.However, Skin color lacks a uniform standard \cite{barrett2023skin,heldreth2024skin}, with classifications ranging from the Fitzpatrick Skin Type Scale (6 categories) \cite{fitzpatrick1988validity}, to the Monk Skin Tone Scale (10 categories) \cite{Ellis_2022}, and the L'Oréal Skin Tone Map (168 categories) \cite{L’Oréal2020}, and its definition varies widely. Skin tone exists on a continuous spectrum and is influenced by subjective human perception \cite{groh2022towards}, complicating discrete categorisation. Ethnicity is sometimes used as proxy,  but this is inaccurate: skin tone is not equivalent to ethnicity \cite{bulatao2004understanding}. Additionally, collecting enough training data for all 168 L'Oréal skin tones is impractical.

Image augmentation \cite{ali2024web,ansari2024algorithmic,yuan2022edgemixup} and generative methods \cite{corbin2023assessing,pakzad2022circle} can increase dataset diversity artificially, but they introduce their own biases and raise concerns about the representativeness of augmented and generated data. These methods also increase implementation costs through longer training times and greater resource demands. 

Counterfactual techniques \cite{dash2022evaluating} evaluate fairness by manipulating skin tones in input images and checking if the model's predictions remain unchanged. Yet, these approaches suffer from the same issues as regularization and reweighing. Questions such as “What would the outputs be if the skin tone were lighter?” are highly subjective, and there is no consensus on the baseline tone for these conversions. Ensuring realistic and natural tone conversion adds another layer of difficulty.

Unearning \cite{bevan2022detecting,alvi2018turning,kim2019learning} and adversarial learning \cite{li2021estimating,hwang2020exploiting} methods involve auxiliary classifiers that detect sensitive attributes like skin color while attempting to suppress the model's reliance on them.  However, they still require clearly defined, classifiable categories for skin tone, a fundamentally continuous and diverse attribute that defies neat classifications. 

Federated learning \cite{xu2022achieving,xing2025achieving} aims to balance fairness and data privacy across distributed devices. However, deploying large-scale  Transformer models across clients requires high-speed communication infrastructure and computational resources, making it challenging to implement in real-world environments.

To overcome these limitations and integrate fairness into skin lesion classification, we propose a skin-independent Prune Learning model. By avoiding reliance on skin tone, our approach eliminates associated biases and redirects the model's attention to focus on the lesion features. This facilitates model compression while preserving performance. Specifically, we focus on skewness, which is on convolutional filters, patch embeddings, and attention heads in CNNs and Transformers, respectively. Components that exhibit excessive focus are pruned,  reducing computational load and bias. Our proposed Prune Learning algorithm supports the practical implementation of fairness techniques, promoting equitable AI applications in clinical settings.

\section{Related Work}
Various approaches have been proposed to enhance fairness through techniques such as pruning and early stopping. FairPrune \cite{wu2022fairprune} aims to improve fairness by performing pruning based on saliency differences using the second derivatives of model parameters, applied to mini-batches stratified by sensitive attributes. This method mitigates inter-group bias by removing parameters that contribute disproportionately to accuracy for specific subgroups. 

The Multi-Exit Framework \cite{chiu2023toward} builds on the observation that a decrease in Soft Nearest Neighbor Loss leads to increased bias. As the network deepens, feature representations of sensitive attributes tend to diverge, resulting in fairness degradation. To counter this,  the framework introduced internal classifiers in the intermediate layers, and halts inference early when the classifier's confidence surpasses a certain threshold. 

SCP-FairPrune \cite{kong2024achieving}, is an extension of the Soft Nearest Neighbor Loss (SNNL) method with Multi-Exit framework,  By using SNNL, it identifies and selectively prunes channels contributing to unfairness. Fairness is improved iteratively through repeated pruning and fine-tuning. However, these methods fall within the group fairness paradigm, relying on differences between predefined groups. In experimental practice, sensitive attributes are excluded during interference from inputs, making it unclear whether fairness measurements stem from genuine attribute mitigation or from correlated confounders. It also becomes difficult  to attribute observed disparities in performance to protected attributes like gender or other hidden  factors.

While these methods assume group differences, they do not explore the underlying  causes of these differences, limiting their interpretability in terms of fairness terms. Furthermore, neither FairPrune nor SCP-FairPrune reports the computational cost or the size of the model. Notably, SCP-FairPrune, merely masks unfair channels rather than physically pruning them, which does not reduce model weights and may even increase computational cost. These approaches are optimized specifically for deep CNNs (e.g., VGG architectures) and rely heavily on strong structural assumptions, making them difficult to generalise to architectures like Vision Transformers, which only use convolution in initial layers. 

FairQuantize \cite{guo2024fairquantize} is another method that also seeks to reduce model size while maintaining fairness, but via weight quantization rather than structural pruning. It retains the original model structure and modifies only the numerical representation of weights, differing fundamentally from pruning, which involves removing structural components. 

General pruning methods often degrade performance for underrepresented subgroups. To address this, \cite{meyer2022fair} proposed an extension to standard cross-entropy loss with sample weighting  and soft label learning, using predictions from the original model as traditional training target, to improve subgroup fairness. 

FairGrape \cite{lin2022fairgrape} calculates the importance of each weight per subgroup and performs pruning while preserving relative importance between groups. However, like others, this method assumes access to explicit subgroup labels beyond the image itself and offers no evidence of computational cost benefits.

These earlier studies suffer from similar limitations seen in regularization and reweighting methods: they require feature grouping, complex computations and are largely CNN-specific. The use of Vision Transformers remains underexplored, and the practical efficiency gains of pruning-based fairness have not been sufficiently evaluated. 

In contrast to these approaches, our method avoids focusing directly on skin color and circumvents the need for group-wise fairness assumptions or explicit attribute classification.  The research questions and contributions of this paper are listed below:
\subsection{Research Questions}
\begin{itemize}
    \item \textbf{RQ1:} Does Prune Learning improve the fairness of skin lesion classification models by removing latent skin features?
    \item \textbf{RQ2:} Can Prune Learning preserve classification performance while improving fairness?
    \item \textbf{RQ3:} Can Prune Learning reduce computational costs while improving fairness?
\end{itemize}

\subsection{Contribution}
\begin{itemize}
    \item Prune Learning for Fairness mitigates bias related to skin color without relying on the complex calculations required by many traditional methods.
    \item The approach effectively prunes irrelevant features while preserving key lesion-related features, maintaining predictive accuracy and avoiding the typical trade-off between fairness and performance.
    \item It significantly reduces the number of model parameters and computational cost while preserving both fairness and diagnostic reliability.
    
\end{itemize}

\section{Methodology}
\subsection{CNN-based Models}
Our objective is to ensure that the model diagnoses skin lesions fairly, independent of the patient's skin color. To achieve this, we identify and remove convolutional channels that disproportionately focus on skin tone features, thereby mitigating their influence on model's classification outcomes?. For CNN-based models, channel selection is performed using the output of convolutional layers followed by  ReLU activation, with max pooling applied to visualize regions of strong model response. Max pooling highlights areas where the model exhibits the highest activation, helping us to determine which parts of the image, lesion or skin, are being emphasized.  We analyse the distribution of feature importance values using skewness, a statistical measure that indicates the asymmetry of a data distribution, specifically the direction and extent of the distribution's tails. Skewness reveals whether a channel is focused on localized or widespread regions:
\begin{itemize}
    \item \textbf{Positive skewness}: The distribution has a long tail on the right side, indicating high activation values in a small localized region, typically a lesion.
    \item \textbf{Zero skewness}: The distribution is symmetrical, indicating no bias toward a specific area. This indicates a channel with little information.
    \item \textbf{Negative Skewness}: The distribution has a long left tail, indicating moderate activation spread across a large area, typically the skin.
\end{itemize}
\begin{figure}
    \centering
    \includegraphics[width=\linewidth]{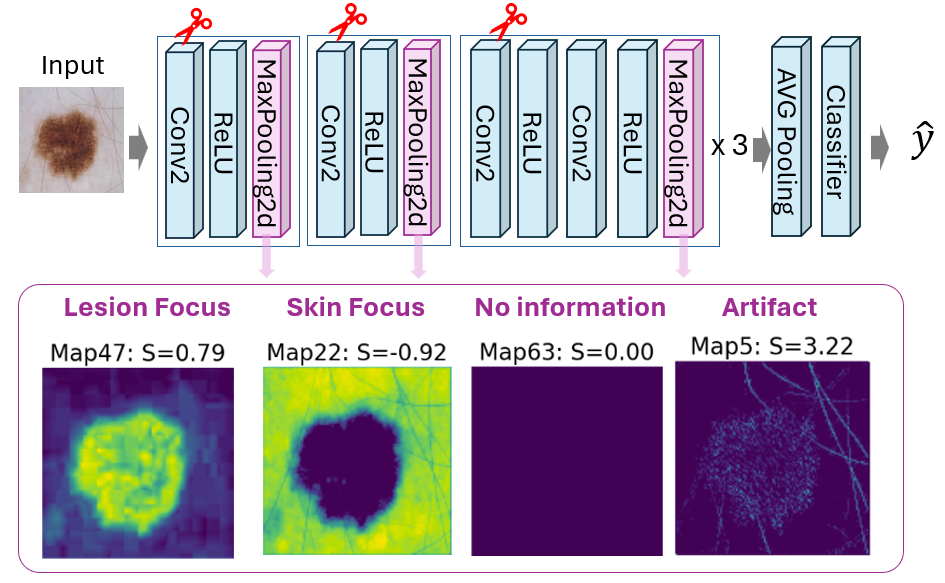}
    \caption{Pruning Process and Target for VGG}
    \label{fig:process_vgg}
\end{figure}
In medical images like skin lesion images, unlike general images, such as those in ImageNet \cite{deng2009imagenet}, lesions are limited to a small part of the image \cite{yoon2024domain,guan2021domain}. Therefore, channels that focus on lesions tend to have low activation in most regions,  and high activation in a small lesion area, resulting in a positively skewed distribution. On the other hand, since the skin occupies the majority of the image, skin-focused channels show more uniformly moderate with only small dips near lesions, resulting in a negatively skewed distribution.

Our pruning method reduces model size and computational cost by not merely masking channels, but by removing entire filters that are deemed unnecessary, which corresponds to the Negative Skewness of the median of the feature map distributions. Specifically, we apply the algorithm \ref{alg:cnn_select_keep_channels} to each max pooling layer, which identifies and retains only the most informative channels, those contributing meaningfully to lesion detection. 

\begin{algorithm}
\caption{CNN: Select Keep Channels}
\label{alg:cnn_select_keep_channels}
\begin{algorithmic}
\State \textbf{Output: Channels to keep $K$} 
\State Load model $f(x)$ with trained weights
\State Initialize $Skew$ to store all filter skewness scores of all images
\For{Each input $x$ in Validation set $X$}
    \State Initialize feature map list $M$
    \State Compute feature map $M = f(x)$
    \For{Each input $m$ in Feature Maps of a Max Pooling Layer $M$}
        \State Initialize empty skewness list $S$
        \State Flatten $m$
        \State Compute skewness $s$ of $m$
        \State Append $s$ to $S$
    \EndFor
    \State Append $Skew$ to $S_{all}$
\EndFor
\State Initialize keep index list $K \gets \emptyset$
\For{Each filter index $i$ until length of $Skew$}
    \State Calculate the Median of $Skew_{i}$
    \If {The median $>0$}
        \State Append $i$ to $K$
    \EndIf
\EndFor
\State \textbf{return} Channels to keep $K$
\end{algorithmic}
\end{algorithm}

Next, we perform the actual pruning using the algorithm \ref{alg:cnn_prune}. The pruning target is the convolution layer immediately preceding the max pooling layer. When multiple convolution layers exist upstream of the max pooling layer, filters with matching indices across these layers are pruned simultaneously, enabling a consistent structural reduction. Finally, since pruning alters the number of output channels in the feature extractor, the input feature size to the sequential classification layer is adjusted according to the latter steps of the Algorithm \ref{alg:cnn_prune}.

\begin{algorithm}
\caption{CNN: Prune Filters}
\label{alg:cnn_prune}
\begin{algorithmic}
\State \textbf{Input:} Trained model $f$, Target layer indices $T$, Keep indices per layer $K$
\State \textbf{Output:} Pruned model $f'$
\State Create a new model $f'$ as a copy of $f$
\For {Each layer $t$ from $T$}
    \If{$f'_{t}$ is Conv2D layer}
        \State Get keep indices of $K_{t}$
        \State Create a new Conv2D layer $l_{new}$ with:
        \State - Same in\_channels, kernel\_size, stride, padding as $f'_{t}$
        \State - Out\_channel as length of $K_{t}$
        \State Copy weight and bias from $f'_{t}$:
        \State - $w_{new} \gets w_{old}[K_t]$ of $f$
        \State - $b_{new} \gets b_{old}[K_t]$ of $f$
        \State Replace $t$ with $l_{new}$
    \EndIf
\EndFor
\State Create a new sequential classifier $S_{new}$
\State Compute new in\_features = $|K_{last}| \times H \times W$
\State Copy weight from $S_{old}$:
\State - $w_{new} \gets w_{old}$ of $S_{old}$
\State Replace classifier in $S_{new}$ with $S_{old}$
\State \textbf{return} Pruned model $f'$
\end{algorithmic}
\end{algorithm}

\subsection{Transformer-based Models}
CNN-based models and ViT differ fundamentally in both model structure and computational complexity. As such, pruning techniques developed for traditional CNNs cannot be directly applied to ViTs. However, in ViTs, certain image patches and attention heads may exhibit overreliance on features such as skin tone, making it possible to identify and selectively prune them.  Prior research \cite{liang2022not} \cite{naseer2021intriguing} has demonstrated that not all patches are beneficial to prediction. Similarly, studies in NLP \cite{voita2019analyzing} have demonstrated that many Attention Heads are redundant and can be pruned without significantly affecting model accuracy.

\subsubsection{Patch Pruning}
Our first pruning target is the convolution channels in the initial layers of ViT-B16. ViT-B16 processes input images by dividing them into 16×16 patches. Let the input image be $x\in \mathbb{R}^{H\cdot W\cdot C}$, where $H$, $W$, and $C$ denote height, width, and channels, respectively. Let $P$ denote patch resolution, and $N=\frac{H\cdot W}{P^{2}}$ be the number of patches. Then, the flattened 2D patch becomes $x_{p}\in \mathbb{R}^{N\times (P^{2}\cdot C)}$ , following the notation of  \cite{dosovitskiy2020image}). 

In our method, we first extract features using a convolution layer prior to patch division. From this layer, we retain only those filters that capture skin-color-independent features using algorithm \ref{alg:cnn_select_keep_channels}. Let $K\subseteq \left\{0, 1, ..., N-1  \right\}$ denote the indices of channels to keep. Applying $K$ to the input image $x$ yields  $x^{keep}\in \mathbb{R}^{H_{K}\cdot W_{K}\cdot C_{K}}$, which is then divided into patches using \ref{eq:x_keep}.

\begin{equation}
\label{eq:x_keep}
    x_{p}^{keep}=\left\{ x_{p}^{(i)} | i\in K \right\}\in \mathbb{R}^{K\cdot (P^{2}\cdot C)}
\end{equation}

\begin{figure}
    \centering
    \includegraphics[width=1\linewidth]{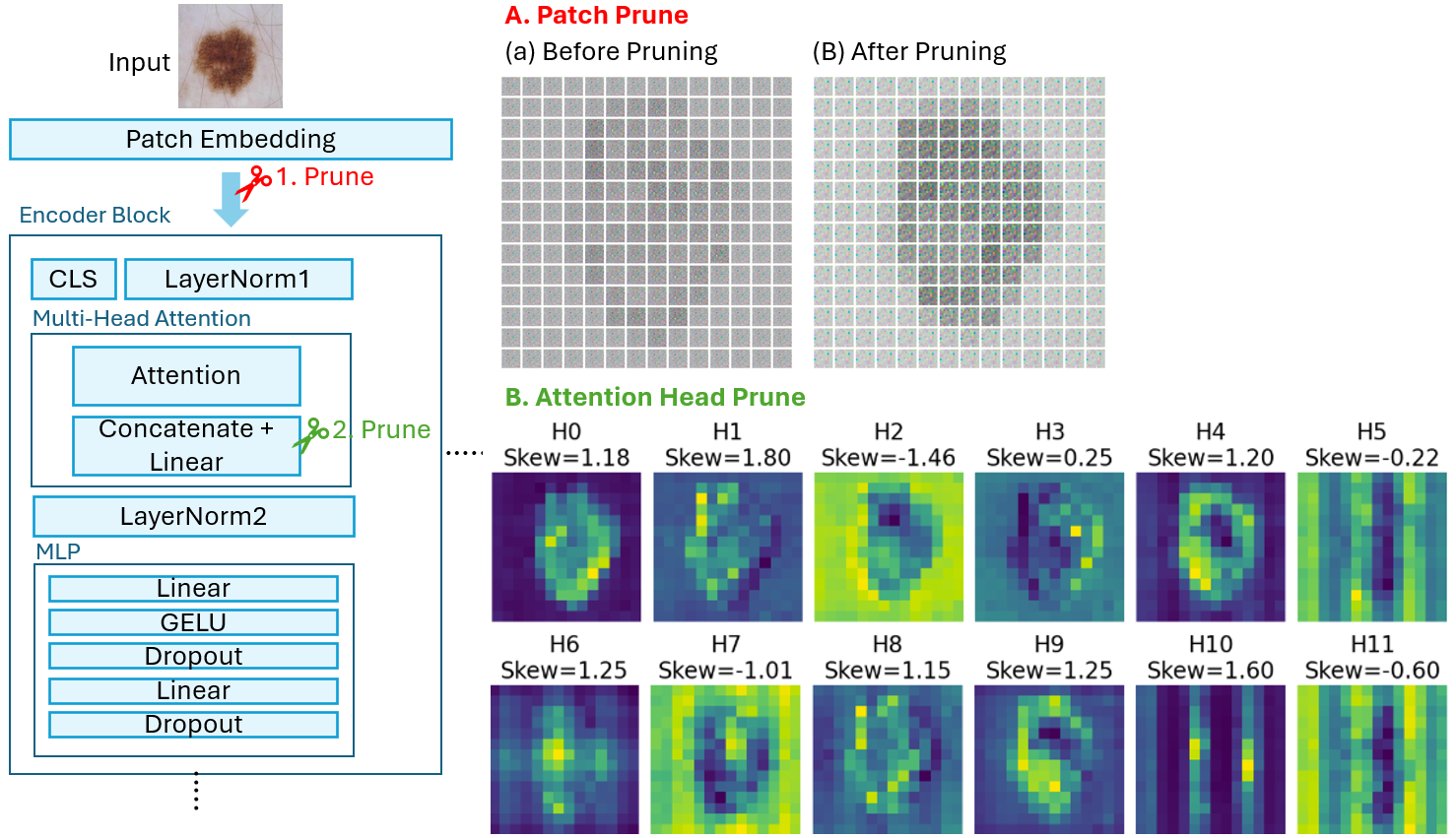}
    \caption{Pruning Process and Target for ViT-B16}
    \label{fig:process_vit}
\end{figure}

Only the channels in $K$ are included in these patches. Fig. \ref{fig:process_vit} A-(a) shows the output using standard 16×16 patching without pruning. In contrast,  Fig. \ref{fig:process_vit} A-(b) shows the output after removing skin-dependent channels (identified via Negative Skewness), reducing resolution to 10×10 based on the greatest common divisor. Comparing the two, reveals a clear emphasis on lesion areas in the pruned version.

The set $K$ is propagated into the first encoder block. The self-attention mechanism is then adapted  to operate only over these retained dimensions. 

\subsubsection{Head Pruning}
While  Patch Pruning removes input patches with strong skin color dependency, attention heads within the self-attention mechanism may still reflect such bias. This is because patches and attention heads have distinct roles: the former influences input representations, while the latter governs how relationships are encoded in the latent space. Consequently, we perform Head Pruning to remove heads that overemphasize skin color.

To do this, we use the statistical skewness of attention maps across samples. We retain only those heads whose attention distributions exhibit positive skewness.  Let  $h\subseteq \left\{\text{head}_{0}, \text{head}_{1}, ..., \text{head}_{H-1}  \right\}$ represent the selected subset, where $H$ is the total number of Heads (similar to channel selection in Algorithm \ref{alg:cnn_select_keep_channels} except that feature maps are replaced with attention maps). The attention is computed using the standard formulation from \cite{vaswani2017attention}, where $Q$, $K$, $V$, and $k$ are query, key, value, and index of key:

\begin{equation}
\label{eq:head_attention}
    \text{Attention}(Q, K, V) = \text{softmax}\left( \frac{Q\cdot K^{\top }}{\sqrt{d_{k}}} \right)V
\end{equation}

Since attentions are calculated using Eq. \ref{eq:head_attention}, attention maps are formulated Eq \ref{eq:attentionmap} below to observe the skewness. The heads that attention maps showed negative skewness were pruned, and then the equations below were performed.

\begin{equation}
\label{eq:attentionmap}
    \text{AttentionMap}(Q, K) = \text{softmax}\left( \frac{Q\cdot K^{\top }}{\sqrt{d_{k}}} \right)
\end{equation}

The outputs for the selected heads are concatenated and converted into the final output presentation using \ref{eq:head_concat}.

\begin{equation}
\label{eq:head_concat}
\text{MultiHead}_{pr} = \text{Concat}\left( \left\{ h_{i}| i\in H \right\} \right)W^{o_{h}}
\end{equation}
Where
\begin{equation}
h_{i}=\text{Attention}\left( QW^{Q}_{i}, KW^{K}_{i}, VW^{V}_{i} \right)
\end{equation}

While some dimensions are deleted by Pruning, the subsequent layers still have a structure that assumes the original number of dimensions, so consistency must be maintained. We ensure this via Algorithm  \ref{alg:vit_residual} which preserves architectural integrity while filtering out irrelevant features.

\begin{algorithm}
\caption{ViT: Residual Add with Pruned Dimension}
\label{alg:vit_residual}
\begin{algorithmic}
\State \textbf{Input:} Input Feature $x_{in} \in \mathbb{R}^{(b\times t \times |K|)}$ where $b$, $t$ and $K$ are batch size, the number of tokens, and the set of index to keep of previous layer
\State \textbf{Output:} Output Feature $x_{out} \in \mathbb{R}^{(b\times t \times d)}$ where $d$ is the output dimension
\If{$d > |K|$}
    \State Initialize zero-padded tensor $x\acute{}_{out} \in \mathbb{R}^{(b\times t \times d)}$
    \For
        \State Set $x\acute{}_{out}(b, t, k)$ with $x_{in}(b, t, I)$ $k=k_{i}\in K$
    \EndFor
    \State $x_{out} \gets x_{out} + x\acute{}_{out}$
    \Else
    \State $x_{out} \gets x_{out} + x_{in}$
\EndIf
\State \textbf{return} $x_{out}$
\end{algorithmic}
\end{algorithm}

\subsection{Evaluation}
\label{sec:evalution}
We evaluate the proposed method by addressing the three research questions (RQs), each corresponding to a key perspective:

\textbf{RQ1: Fairness Improvement}
We assess whether the fairness of the model can be improved by pruning filters related to skin color. Our method reduces outputs related to skin characteristics in the latent space, enabling fairer predictions that are independent of skin tone. In contrast, baseline methods for comparison operate only on categorical sensitive attributes. As a result, we adopt a group fairness framework, classifying skin color into two groups, Light and Dark. Fairness is quantitatively evaluated using three metrics inspired by \cite{wu2022fairprune}:  
\begin{itemize}
    \item True Negative Rate Equal Opportunity (EOpp0) Eq.\ref{eq:eopp0}
    \item True Positive Rate Equal Opportunity (EOpp1) Eq. \ref{eq:eopp1}
    \item Equalized Odds (EOdd) \ref{eq:eodd}
\end{itemize}
These metrics assess the similarity of classification performance across groups.

EOpp0 measures the average absolute difference in True Negative Rate (TNR) between groups across all classes: 

\begin{equation}
\label{eq:eopp0}
\text{EOpp0}=\frac{1}{K}\sum_{k=1}^{K}|TNR^{(k)}_{s=0} -TNR^{(k)}_{s=1}|
\end{equation}

TNR is defined using the One-vs-Rest strategy, where a class $k$ is considered positive and all other classes are considered negative. The TNR for class $k$ is evaluated as the proportion of samples that were correctly predicted as not belonging to class $k$, as shown in Equation \ref{eq:tnr}. Additionally, $\varepsilon = 1e-9$ is added to the denominator to prevent division by zero in minority classes,

\begin{equation}
\label{eq:tnr}
\text{TNR}=\frac{\text{True Negative}}{\text{True Negative}+\text{False Positive} + \varepsilon}
\end{equation}

\begin{equation}
\label{eq:eopp1}
\text{EOpp1}=\frac{1}{K}\sum_{k=1}^{K}|TPR^{(k)}_{s=0} -TPR^{(k)}_{s=1}|
\end{equation}

\begin{equation}
\label{eq:tpr}
\text{TPR}=\frac{\text{True Positive}}{\text{True Positive}+\text{False Negative} + \varepsilon}
\end{equation}

\begin{align}
\label{eq:eodd}
\text{EOdd} = \frac{1}{K} \sum_{k=1}^{K} \big( 
& \left| \text{TPR}^{(k)}_{s=0} - \text{TPR}^{(k)}_{s=1} \right| \nonumber \\
& + \left| \text{FPR}^{(k)}_{s=0} - \text{FPR}^{(k)}_{s=1} \right| \big)
\end{align}

\begin{equation}
\label{eq:fpr}
\text{FPR}=\frac{\text{False Positive}}{\text{False Positive}+\text{True Negative} + \varepsilon}
\end{equation}

\textbf{RQ2: Predictive Performance Maintenance}
We examine whether pruning for fairness compromises predictive performance. Metrics include Precision, Recall, and F1-Score. Results from the proposed method are compared with those from the unpruned (vanilla) model to determine if fairness gains come at the cost of accuracy.

\textbf{RQ3: Computation Cost Reduction}
We evaluate reductions in computational cost by comparing our model to the vanilla model using the following metrics: 

\begin{itemize}
    \item Total number of model parameters
    \item Floating Point Operations per Second (FLOPs)
    \item Memory Footprint (MF)
\end{itemize}

These are benchmark indicators for efficiency in prune learning. Additionally, we assess the number of training epochs needed to reach the peak F1-score, offering insights into training efficiency. This allows us to measure not only inference savings but also improvements in learning dynamics.

\section{Experiment Setup}
\subsection{Dataset}
We used the ISIC2019 \cite{tschandl2018ham10000,codella2018skin,hernandez2024bcn20000}, a benchmark widely adopted in skin lesion classification and fairness research. To compare with SCP-Prune Fairness, we applied the same train/validation/test split published on GitHub by \cite{kong2024achieving}. 

The dataset includes skin color labels based on the Fitzpatrick skin type (1–6). We  grouped these into two categories for evaluation purposes:  types 1–3 as "light skin" and types 4–6 as "dark skin". This grouping is an intentional processing step designed to ensure a comparison with other methods. However, it is important to note that our method is designed to eliminate overreliance on skin features, making it inherently independent of explicit skin tone labels or groupings. Thus, in contrast to conventional methods, our approach does not require these labels for training or evaluation.

\subsection{Models}
We evaluated our approach using two representative models. 
\begin{itemize}
    \item VGG11: A CNN model selected due to its use in the FairPrune baseline. Training was performed for up to 200 epochs, and the model with the highest F1 score was selected as the vanilla model. We used Stochastic Gradient Descent (SGD) as the optimizer, with a ReduceLROnPlateau rate scheduler. The initial learning rate was set to 1e-2, with a patience value of 10, and a decay factor of 0.5. Details of the VGG11 model are shown in the “Vanilla” column of Table \ref{tab:results}.

    \item ViT-16B: A transformer-based model that has recently become state-of-the-art in skin lesion classification tasks due to its high performance. Similar to VCG11, the model checkpoint with the highest F1 score was selected as the vanilla model. A cosine scheduler with a warm-up period was employed, and AdamW was used as the optimizer, as it is well-suited for transformers. The initial learning rate was set to 1e-5, and the maximum number of epochs was set to 100. The performance of the ViT-16B vanilla model is shown in Table \ref{tab:results}.
\end{itemize}

We applied our fairness pruning methods to these vanilla models and fine-tuned the pruned models using the same settings as their vanilla counterparts. Evaluation followed the metrics defined in section \ref{sec:evalution}. 

For ViT-B16, we tested six pruning configurations as shown in Table \ref{tab:pruning_pattern}. The configurations "Prune1" and "Prune2" correspond to the pruning locations shown in Fig. \ref{fig:process_vit}. 
\begin{itemize}
    \item Pattern 1: No fine-tuning (vanilla model).
    \item Patterns 2–5, Full re-training performed after pruning.
    \item Pattern 6: Patch pruning followed by initial training, then head pruning with re-training; during this final stage, the weights of the patch embedding layer are frozen.
\end{itemize}

This systematic comparison allows us to assess both the performance and the fairness impact of different pruning strategies.
\begin{table*}[]
\centering
\begin{tabular}{l|cc|c|c}
\hline
\multicolumn{1}{c|}{\multirow{2}{*}{Pruning Pattern}} & \multicolumn{2}{c|}{Patch Index Base} & Attention Head Base & \multirow{2}{*}{Fine-Turning} \\
\multicolumn{1}{c|}{}                                 & Prune 1           & Prune 2           & Prune 2             &                               \\ \hline
1 .Vanilla                                            & N                 & N                 & N                   & No                            \\
2. Skew Prune (Patch)                                 & Y                 & N                 & N                   & Full                          \\
3. Skew Prune (Patch)                                 & Y                 & Y                 & N                   & Full                          \\
4. Skew Prune (Head)                                  & N                 & N                 & Y                   & Full                          \\
5. Skew Prune (Patch+Head)                            & Y                 & N                 & Y                   & Full                          \\
6. Skew Prune (Patch+Head)                            & Y                 & N                 & Y                   & Partial                       \\ \hline
\end{tabular}
\caption{ViT: Pruning Patterns}
\label{tab:pruning_pattern}
\end{table*}

\section{Results}
\subsection{VGG11 Evaluation}

The experimental results for the VGG11 model are summarized in Table \ref{tab:results}. 

\textbf{Vanilla Model (Column 1)}

The baseline model achieved accuracy of 78\%, and an F1 score of 66\%, which is consistent with performance levels typically reported in fairness studies involving CNN-based models.  In terms of fairness metrics, 

\begin{itemize}
    \item EOpp0 (True Negative Rate gap) was favorable at 1\%,
    \item EOpp1 (True Positive Rate gap) showed a notable group disparity of over 6.5\%
    \item EOdd (Equalised Odds) indicated a group difference of approximately 8\%.
\end{itemize}

The computational cost was also standard, with no notable excess or deficiency.

\textbf{SCP-FairPrune (Column 2)}   

Following the methodology of  \textbackslash{}cite\{kong2024achieving\}, we applied  three rounds of fine-tuning and a 2\% pruning rate on the last feature layer. Results from the third iteration are reported: With this method, 

Accuracy and  F1 score improved by 2\% and 1\% compared to the vanilla model. 

\begin{itemize}
    \item EOpp0 improved to below 1\%,
    \item EOpp1 deteriorated slightly, through the increase was less than 1\%.
    \item EOdd was significantly reduced to approximately 0.3\%.
\end{itemize}

However, since SCP-FAirPrune is not designed for lightweighting, there was no reduction in FLOPS or parameter count. In fact, the memory footprint increased due to masking overhead. Notably, only 30 filters were removed after three pruning iterations.

 \textbf{SkewPrune (Our method, Column 5) }
 
 SkewPrune identified and pruned 229 filters associated with skin-color dependency, including the corresponding output channels. Improvements were observed across all fairness metrics:
 
  \begin{itemize}
      \item EOpp0 showed a slight enhancement
      \item EOpp1 improved by over 1.5\% from the vanilla model to 5.11\%,
      \item EOdd improved by 1.6\%.
  \end{itemize}
  
  Crucially, predictive performance was maintained with an accuracy at 79\% and an F1 score of 65\%, indicating no performance trade-off for fairness gains. 
  
  In terms of efficiency:
 
\begin{itemize}
    \item FLOPS per image decreased from 7.61 to 6.99
    \item Model parameters were reduced to 107.9M,
    \item Memory footprint was reduced by approximately 80 MB.
\end{itemize}

Thus, SkewPrune achieved a balanced optimization, improving fairness, maintaining predictive accuracy, and reducing computational cost.

\textbf{Additional Comparison: SCP-FairPrune with 229 pruned channels (Column 3)}

To further compare the pruning impact, we applied SCP-FaiPrune with 229 channels pruned. Fine-tuning was performed once. The results showed: 

\begin{itemize}
    \item EOpp0 was similar to the vanilla model
    \item EOpp1 and EOdd improved by over 1\%, but not to the same extent as SkewPrune.
    \item Accuracy dropped by approximately 2\%
    \item Computational cost remained unchanged
\end{itemize}

Moreover, the model required 159 epochs to reach the highest F1 score, compared to 116 epochs for SkewPrune, revealing slower convergence by 43 epochs.

\textbf{Additional Comparison: SCP-FairPrune with 229 removed channels (Column 4)}

Similar to our method, this not only masked 229 channels during pruning, but also removed the filter structure.

\begin{itemize}
    \item EOpp0, bias was mitigated compared to the vanilla model
    \item EOpp1 and EOdd degraded approx. 3\%
    \item Accuracy was 1\% higher than vanilla
    \item Computational cost remained the vanilla model due to reducing masking information.
\end{itemize}

In summary, SkewPrune offers a robust trade-off-free solution for improving fairness, maintaining accuracy, and reducing resource demands in CNN-based skin lesion classification.
\begin{table*}[ht]
\centering
\begin{tabular}{ll|r|r|r|r|r}
\hline \hline
\multicolumn{2}{l|}{\textbf{Metrics}} 
& \makecell[r]{1.Vanilla} 
& \makecell[r]{2.SCP-Fair \\Prune} 
& \makecell[r]{3.SCP-Fair \\Prune\\(n=229)} 
& \makecell[r]{4.SCP-Fair \\Prune\\Removed} 
& \makecell[r]{5.SkewPrune} \\ 
\hline
\multirow{2}{*}{\textbf{Performance}}    
& Accuracy & 0.78 & 0.80 & 0.76 & 0.79 & 0.79 \\
& F1-score & 0.66 & 0.67 & 0.65 & 0.66 & 0.65 \\ 
\hline
\multirow{3}{*}{\textbf{Fairness}}       
& EOpp0 & 0.0110 & 0.0051 & 0.0106 & 0.0068 & 0.0099 \\
& EOpp1 & 0.0666 & 0.0691 & 0.0530 & 0.0937 & 0.0511 \\
& EOdd  & 0.0776 & 0.0742 & 0.0636 & 0.1005 & 0.0610 \\ 
\hline
\multirow{4}{*}{\textbf{Computational Cost}} 
& FLOPs (G) & 7.61 & 7.61 & 7.61 & 7.61 & 6.99 \\
& Params (M) & 128.8 & 128.8 & 128.8 & 128.8 & 107.91 \\
& Memory (M) & 491.33 & 500.33 & 500.33 & 491.33 & 411.65 \\ 
& Best Epochs & 176 & 82 & 159 & 198 & 116 \\
\hline \hline
\end{tabular}
\caption{VGG11: Evaluation results of Fairness, Performance, and Sustainability Metrics}
\label{tab:results}
\end{table*}

\subsection{ViT-B16}
The detailed results of the ViT-B16 experiments are shown in Table \ref{tab:results_vit}. 
\textbf{Vanilla Model (Column 1) }
The baseline ViT-B16 model, trained without any pruning or fairness interventions, achieved:  
\begin{itemize}
    \item EOpp0 below 0.8\%, indicating a minimal TNR disparity across groups.
    \item However, EOpp1 and EOdd showed significant group differences of approximately 7.4\% and 8.2\%, respectively,  reflecting substantial fairness concerns.
\end{itemize}

Performance was maintained at a high level, while computational costs were within the expected range. 

\textbf{SkewPrune (Patch Pruning Only, Column 2)}
In this configuration, pruning was applied to the Patch Embedding layer, reducing the input dimension from 786 to 336, thereby eliminating over half of the original input channels. This reduced the influence of skin-dependent patches and slightly improved fairness. 
\begin{itemize}
    \item EOpp1 improved to 6.7\%
    \item EOdd improved to 7.8\%.
\end{itemize}

Performance remained stable, and computational cost was significantly reduced due to the complete removal (not just masking) of unused patches. 

\textbf{SkewPrune (Patch Pruning with Early Encoder Head Pruning, Column 3}

This pattern involved pruning from the patch embedding stage to the head of the first encoder block. The results showed a decline in fairness. 

\begin{itemize}
    \item EOpp1 and EOdd worsened by 2.5\% and 1.8\%, respectively
    \item Performance metrics remained similar.
\end{itemize}

The drop in fairness is attributed to pruning heads prematurely, which disrupted latent feature representations. However, computational savings were observed due to the simultaneous removal of related attention heads.

\textbf{SkewPrune (Head Pruning Only, Column 4)}

This configuration focused on pruning attention heads with high dependency on skin color, based on skewness analysis of attention maps. 

\begin{itemize}
    \item Fairness improved by 1\% across all metrics.
    \item Accuracy and F1 scores increased by about 1\% and 2\% over the vanilla model.
\end{itemize}

In addition to pruning queries, keys, and values, the associated normalization layers were also removed. This not only eliminated skin-biased attention but also reinforced focus on lesion-relevant features, mimicking guided learning and enhancing classification performance. 

\textbf{SkewPrune (Patch + Head Pruning with Full Fine-Tuning, Column 5)}

In this experiment, both patch and head pruning were performed, followed by fine-tuning. Compared to the vanilla model: 

\begin{itemize}
    \item EOpp1 improved by 0.28\%
    \item EOdd improved by 0.12\%
\end{itemize}

However, these gains were less pronounced than the head-only pruning case. This is attributed to the re-training process updating the initially fair patch embedding weights, diminishing the earlier fairness advantages. 

\textbf{SkewPrune (Patch + Head Pruning with Frozen Embeddings, Column 6)}

In this final configuration, patch embedding learned during patch pruning was frozen, and only encoder blocks were fine-tuned after head pruning. This approach leveraged the synergistic effects of Patch Prune and Head Prune, achieving the best fairness outcomes. Compared to the vanilla model: 
\begin{itemize}
    \item EOpp1 improved by 1.18\%
    \item EOdd by 1.11\%
    \item The disparity in EOpp1 between groups fell below 6.3\%
\end{itemize}

Efficiency gains included:

\begin{itemize}
    \item 0.07 GFLOPs reduction
    \item 0.33M fewer parameters
    \item 7.13M lower memory footprint
\end{itemize}

Crucially,  these improvements were realized without sacrificing performance, and computational efficiency was improved. This demonstrates that combining patch and head pruning, when carefully staged and controlled, can balance fairness, accuracy, and resource efficiency in ViT-based skin lesion classification.
\begin{table*}[]
\centering
\begin{tabular}{ll|r|r|r|r|r|r}
\hline \hline
\multicolumn{2}{l|}{Metrics}  & 1. Vanilla & \begin{tabular}[c]{@{}l@{}}2. Skew\\      Prune\\      (Patch)\end{tabular} & \begin{tabular}[c]{@{}l@{}}3. Skew\\      Prune\\      (Patch)\end{tabular} & \begin{tabular}[c]{@{}l@{}}4. Skew\\      Prune \\      (Head)\end{tabular} & \begin{tabular}[c]{@{}l@{}}5. Skew\\      Prune\\       (P+H)\end{tabular} & \begin{tabular}[c]{@{}l@{}}6. Skew\\      Prune\\       (P+H)\end{tabular} \\ \hline
\multirow{2}{*}{Performance}    & Accuracy & 0.82 & 0.82 & 0.83 & 0.83 & 0.83 & 0.83 \\
                                & F1-score & 0.73 & 0.72 & 0.73 & 0.75 & 0.74 & 0.73 \\ \hline
\multirow{3}{*}{Fairness}       & EOpp0 & 0.0080 & 0.0110 & 0.0099 & 0.0073 & 0.0096 & 0.0086 \\
                                & EOpp1 & 0.0742 & 0.0678 & 0.0995 & 0.0645 & 0.0714 & 0.0624 \\
                                & Eodd & 0.0822 & 0.0788 & 0.1095 & 0.0718  & 0.0810 & 0.0711 \\ \hline
\multirow{4}{*}{Computing Cost} & GFLOPs & 11.29 & 11.22 & 10.96 & 11.29 & 11.22 & 11.22 \\
                                & Params (M) & 57.3 & 56.97 & 55.64 & 57.3 & 56.97 & 56.97\\
                                & Memory Footprint(M) & 327.3  & 321.93 & 313.37 & 325.82 & 320.19 & 320.19 \\
                                & Best Epochs & 87 & 74 & 85 & 67 & 56 & 69 \\ \hline \hline
\end{tabular}
\caption{ViT-B16: Evaluation results of Fairness, Performance, and Sustainability Metrics}
\label{tab:results_vit}
\end{table*}

\section{Discussion}
\subsection{Fairness through Skewness-Based Pruning }
The experimental results from both VGG and ViT-B16 demonstrate that pruning internal structures associated with skin color bias, guided by skewness in the latent semantic space, is an effective approach for improving fairness. In particular, measuring skewness in feature distributions offers a concise and statistically grounded technique, especially suited to the homogenous visual nature of skin lesion images.

Importantly, this method eliminates the need for explicit skin color classification, annotation, or complex preprocessing typically required by conventional fairness techniques. In the VGG model, fairness improvements were achieved with simple steps, while predictive performance and computational cost were reduced. In ViT-B16, the effects of Patch Pruning and Head Pruning were validated both independently and in combination. The synergistic effect of their integration was evident.

As expected from theory, removing skin-focused structures implicitly guided the model's attention toward clinically relevant lesion areas. Thus, fairness gains were obtained without compromising performance. Additionally, pruning led to tangible reductions in computational cost, proportional to the extent of structure removed.

Overall, the study shows that the three research questions can be answered positively: prune Learning can improve the fairness of
skin lesion classification models by removing latent skin features (RQ1). At the same time, it can preserve classification performance (RQ2) and simultaneously reduce computational costs (RQ3).

Although this study is based on publicly available datasets, the proposed method addresses a critical barrier in the clinical adoption of AI for dermatology: ensuring equitable diagnostic performance across diverse patient populations. Skin tone has been shown to influence the accuracy of lesion classification models, potentially leading to underdiagnosis or misdiagnosis in patients with darker skin. Our method improves fairness without requiring skin color annotations, making it practical for clinical settings where such metadata may be unavailable or ethically sensitive. Moreover, by reducing computational costs and model size, the approach supports deployment on edge devices, enabling point-of-care diagnostics in low-resource environments. These properties make the method directly applicable to real-world clinical workflows.

\subsection{Research Limitations}
Although the pruning strategy in this study was designed primarily to address fairness, it also brought about improvements in model compactness and computational efficiency. These benefits contribute to reducing the resource and computational costs that raise barriers to the practical implementation of fairness methods in lesion detection. However, this work did not fully address general-purpose model pruning, namely reducing redundancy in large-scale pre-trained models, such as those trained on ImageNet. Such models often include a surplus of filters or attention heads that are irrelevant to specific tasks like skin lesion classification. Further optimization opportunities remain in this direction.

\subsection{Future Work}
Our findings reveal that models,  both CNNs and ViTs, that have been pretrained on general-purpose images such as ImageNet, contain unnecessary components when adapted to medical imaging tasks characterised by a small number of classes and high visual consistency. In future work, we plan to: 
\begin{itemize}
    \item Extend the pruning approach to other medical image classification tasks
    \item Explore pruning as a general framework for task-specific model compression, and
    \item Implement these lightweight models on edge devices and validate their practicality in federated learning environments.
\end{itemize}

\section{Conclusion}
As AI becomes increasingly prevalent in life-critical fields such as healthcare, concerns about systemic bias, such as that introduced by skin color, are growing. Conventional methods to mitigate this bias have struggled due to the complexity of skin tone as a sensitive attribute, as well as challenges in quantitative fairness evaluation, algorithmic complexity, and performance trade-offs.

In this study, we proposed a novel fairness-enhancing method based on detecting distribution skewness in the feature space. By pruning filters, patches, and attention heads associated with skin-color bias, we demonstrate a way to improve fairness while simultaneously reducing computational requirements. Notably, our application of pruning to both patch embeddings and attention heads in ViT-B16, from a fairness perspective, represents a novel contribution to the literature.

This method eliminates the need for explicit attribute labels or sensitive-group classification and reduces dependence on cumbersome preprocessing. It therefore offers a practical and scalable solution to fairness challenges in skin-color-sensitive medical AI. Ultimately, our approach marks a meaningful step toward developing more equitable, efficient, and deployable healthcare technologies.

\section*{Acknowledgment}
The authors would like to thank the Data Science, Artificial Intelligence, and Modelling (DAIM) Institute at the University of Hull for their support. Furthermore, the authors extend their heartfelt gratitude to Dr. Jun-ya Norimatsu at ALINEAR Corp. for technical advice with the experiments.

\section*{Reference}
\bibliographystyle{IEEEtran}
\bibliography{reference}

\begin{IEEEbiographynophoto}{Kuniko Paxton} is a PhD candidate in Computer Science at the University of Hull, and the funding institution is DAIM. The research interests are Pruning Learning, Fairness in AI, explainability, and Adversarial attacks. Contact her at k.azuma-2021@hull.ac.uk
\end{IEEEbiographynophoto}

\begin{IEEEbiographynophoto}{Koorosh Aslansefat} is an assistant professor of computer science at the University of Hull, HU6 7RX Hull, U.K., affiliated with the Dependable Intelligent System Group. His research interests span artificial intelligence safety, Markov modelling, and real-time dependability analysis. Aslansefat received his PhD in computer science from the University of Hull. He is a Member of IEEE. Contact him at K.Aslansefat@hull.ac.uk
\end{IEEEbiographynophoto}

\begin{IEEEbiographynophoto}{Dhavalkumar Thakker} is a Professor of Artificial Intelligence (AI) and the Internet of Things (IoT) at the University of Hull, where he leads a group focused on Responsible Artificial Intelligence. His research emphasizes AI Explainability, AI Safety, and Fairness. With nearly two decades of experience, Dhaval has been at the forefront of innovative solutions through funded projects. His interdisciplinary research spans Generative AI and the applications of Edge computing alongside IoT technologies.  He has a track record in leveraging AI for Social Good, notably in Smart Cities, Digital Health, and the Circular Economy. Contact him at D.Thakker@hull.ac.uk
\end{IEEEbiographynophoto}

\begin{IEEEbiographynophoto}{Yiannis Papadopoulos} is Professor of Computer Science and Leader of the Dependable Intelligent Systems (DEIS) Research Group at the University of Hull in the U.K. For over 30 years, Papadopoulos and his research group have pioneered cutting-edge model-based, bio-inspired and statistical technologies for the analysis and design of dependable engineering systems, with a recent intense focus and contributions towards achieving trustworthy, safe AI. Many of the software tools that they have crafted, including HiP-HOPS and EAST-ADL, have become commercial and are used in transport and other industries. Contact him at Y.I.Papadopoulos@hull.ac.uk
\end{IEEEbiographynophoto}

\begin{IEEEbiographynophoto}{Tanaya Maslekar} is a medical doctor and University of Birmingham graduate with a strong interest in the intersection of healthcare and technology. Her medical training has deepened her conviction that technology, especially AI can play a pivotal role to improve diagnostic accuracy, patient outcomes, and efficiency of medical workflows. Tanaya collaborates on interdisciplinary projects focused on bridging digital innovation and medicine.
\end{IEEEbiographynophoto}

\end{document}